\documentclass[10pt,twocolumn,letterpaper]{article}
\usepackage{iccv}
\usepackage{times}
\usepackage{epsfig}
\usepackage{graphicx}
\usepackage{blindtext}
\usepackage{amsmath}
\usepackage{amssymb}
\usepackage{multirow}
\usepackage[accsupp]{axessibility}
\usepackage[dvipsnames]{xcolor}

\usepackage{mathtools}

\usepackage[pagebackref=true,breaklinks=true,colorlinks,bookmarks=false]{hyperref}

\usepackage[capitalize,noabbrev]{cleveref}

\newcommand{\vectorname}[1]{{\mathrm{\mathbf{#1}}}}
\newcommand{\myparagraph}[1]{\vspace{4pt}\noindent{\bf #1}}
\newcommand{\heading}[1]{{\bf #1}}

\iccvfinalcopy 

\def\iccvPaperID{33} 

\ificcvfinal\fi

\begin{document}

\title{Actor-agnostic Multi-label Action Recognition with Multi-modal Query}
\author{Anindya Mondal$^1$$^2$$^3$\thanks{Authors have equal contributions.} , Sauradip Nag$^1$$^2$$^5$\footnotemark[1] , Joaquin M Prada$^1$$^4$, Xiatian Zhu$^1$$^2$$^3$, Anjan Dutta$^1$$^2$$^3$$^4$\footnotemark[1] \\
$^1$University of Surrey, $^2$CVSSP, $^3$Surrey Institute for People-Centred AI, \\ $^4$School of Veterinary Medicine, $^5$iFlyTek-Surrey Joint Research Center on AI\\
{\tt\small \{a.mondal, s.nag, j.prada, xiatian.zhu, anjan.dutta\}@surrey.ac.uk,}}

\maketitle
\ificcvfinal\thispagestyle{empty}\fi

\begin{abstract}
Existing action recognition methods are typically {\em actor-specific} due to the intrinsic topological and apparent differences among the actors. This requires actor-specific pose estimation (\eg, humans vs. animals), leading to cumbersome model design complexity and high maintenance costs. Moreover, they often focus on learning the visual modality alone and single-label classification whilst neglecting other available information sources (\eg, class name text) and the concurrent occurrence of multiple actions. To overcome these limitations, we propose a new approach called 'actor-agnostic multi-modal multi-label action recognition,' which offers a unified solution for various types of actors, including humans and animals. We further formulate a novel Multi-modal Semantic Query Network (MSQNet) model in a transformer-based object detection framework (\eg, DETR), characterized by leveraging visual and textual modalities to represent the action classes better. The elimination of actor-specific model designs is a key advantage, as it removes the need for actor pose estimation altogether. Extensive experiments on five publicly available benchmarks show that our MSQNet consistently outperforms the prior arts of actor-specific alternatives on human and animal single- and multi-label action recognition tasks by up to 50\%. Code is made available at \href{https://github.com/mondalanindya/MSQNet}{https://github.com/mondalanindya/MSQNet}.
\end{abstract}

\section{Introduction}
\label{sec:intro}

\begin{figure*}
\centering
\includegraphics[width=\textwidth]{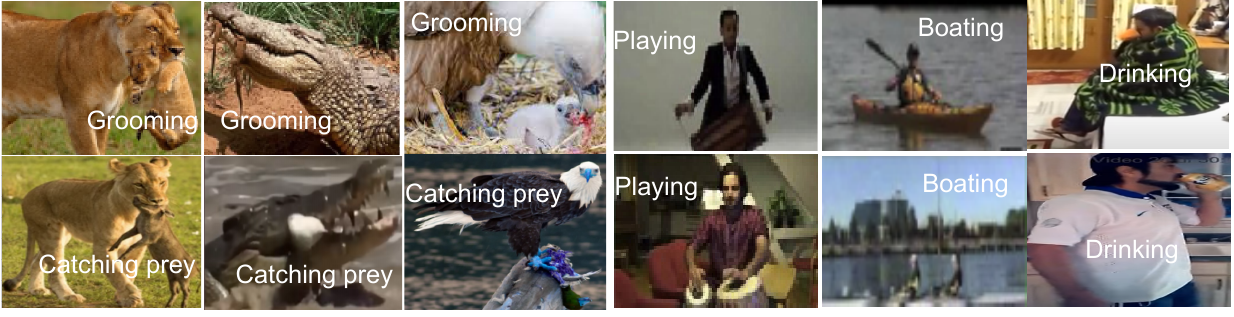}
\caption{
Illustration of large action variation across different actors (\eg, animals and humans). Such differences often motivate the development of actor-specific action recognition models, such as using actor specific pose estimation \cite{ng2022animal}}.
\label{fig:teaser}
\end{figure*}

Action recognition has been extensively studied, focusing on humans as the actors \cite{lea2016segmental,lea2017temporal,farha2019ms,li2020ms,tran2019video,fan2020pyslowfast,kay2017kinetics,arnab2021vivit,bertasius2021space}, benefiting a variety of applications, \eg, 
healthcare \cite{belongie2005monitoring, zhou2020deep}, virtual and augmented reality \cite{li2023action},
and many more \cite{kong2022human}. 
While majority of research has concentrated on humans, there is potential for action recognition to be applied to animals as well \cite{ng2022animal}. However, recognizing actions and behaviors in animals presents a challenging task. Animals often exhibit different shapes, sizes, and appearances, as illustrated in \cref{fig:teaser}. As a result, it becomes necessary to develop more sophisticated and customized designs tailored to the unique characteristics of each animal actor. One approach to achieve this is by incorporating specific pose information of the actors \cite{ng2022animal}. Consequently, the ultimate solution becomes tailored to each specific animal actor.
%
%
%
%
%
Furthermore, most existing methods for action recognition focus on single-label classification, which means they are designed to assign a single action label to a given video. However, in real-world scenarios, multiple actions may occur within a single video, making the task more complex. These methods rely exclusively on video data for model training and inference. As a result, the textual information contained within action class names, often represented as discrete encoded numbers, is often disregarded, even though it could provide valuable context and information.

In order to address the limitations mentioned above, we propose a new problem formulation -- \emph{multi-modal multi-label learning for actor-agnostic action recognition}. This novel problem setting aims to utilize multiple sources of information, including visual and textual data, to predict multiple action labels for each video. An essential prerequisite is for the model to remain independent of the actor's identity, thus ensuring its broader applicability and ease of deployment without relying heavily on specific actor traits. This encourages the advancement of sophisticated action recognition models that offer practical benefits in terms of computation and cost-effectiveness.
In this direction, we introduce a novel Multi-modal Semantic Query Network (MSQNet), inspired by the Transformer based detection framework (\eg, DETR \cite{liu2022dabdetr}). By treating multi-label action classification as a specialized form of object detection, MSQNet eliminates the reliance on explicit localization and actor-specific information like actor's pose, thus making it \emph{actor-agnostic}. Importantly, we design a multi-modal semantic query learning scheme to incorporate visual and textual information using a pretrained vision-language model (\eg, CLIP \cite{gao2021clip}). This approach allows us to combine and utilize both visual and textual data in a trainable manner. This results in a more comprehensive and precise representation of action classes, while simultaneously acquiring the distinct characteristics of actors directly from the training data without requiring actor-specific elements.

Our contributions are summarized as follows: {\bf (1)} We introduce a new problem formulation of {\em multi-modal multi-label learning for actor-agnostic action recognition} in unconstrained videos. Using a single model architecture for various action tasks minimizes the cumbersome need for actor-specific design, improving model generalization and maintenance efficiency. 
{\bf (2)} To tackle this problem, we design a novel {\em Multi-modal Semantic Query Network} (MSQNet) model that casts the multi-label action classification problem into a multi-modal target detection task in the elegant Transformer encoder-decoder framework. It is characterized by a principled vision-language information fusion design for creating richer label queries so that more accurate action classes can be recognized eventually, without requiring actor-specific pose information. 
{\bf (3)} Through rigorous experimentation on five publicly available benchmarks, we demonstrate that our MSQNet consistently surpasses previous more complex, actor-specific alternatives for human and animal multi-label action recognition tasks.

\section{Related Work}

\textbf{Action Recognition:} Accurate encoding of spatial and motion information is crucial for recognizing actions in unconstrained videos. Early attempts at video understanding used a combination of 2D or 3D convolution and sequential models in order to capture the spatial and temporal information \cite{carreira2017quo, feichtenhofer2017spatiotemporal, wang2017spatiotemporal}. Recently, researchers have proposed vision transformer-based models \cite{arnab2021vivit, liu2022video, yan2022multiview}, which effectively consider long-range spatio-temporal relationships and have comfortably surpassed their convolutional counterparts. While earlier models mostly consider uni-modal solutions, recent works, such as ActionCLIP \cite{wang2021actionclip}, and XCLIP \cite{ni2022expanding} adopted multi-modal approach by utilizing CLIP and driving it for video understanding. However, all existing works are \emph{actor-specific}, \ie, they consider actions either by humans \cite{arnab2021vivit,rasheed2022fine} or by animals \cite{ng2022animal}. We aim to address this limitation by addressing an \textit{actor-agnostic} action recognition problem, which, to our knowledge, is the first of its kind.

\textbf{Vision Transformers (ViTs):} Inspired by the success of attention-based Transformer \cite{vaswani2017attention} models in NLP, Dosovitskiy \etal adapted the framework for image classification and named it Vision Transformer (ViT) \cite{dosovitskiy2020image}. 
With the success of ViT, many others came out with their frameworks focusing on efficient training \cite{touvron2020deit}, shifted window-based self-attention \cite{liu2021Swin}, deeper architectures \cite{Touvron2021cait}, self-supervised pretraining \cite{caron2021dino}, etc. Following them, Carion \etal \cite{carion2020end} considered the CNN feature 
maps within a classical Transformer encoder-decoder architecture to design an end-to-end object detection framework called DETR \cite{carion2020end}.
This was improved further by several DETR-like object detectors focusing on training \cite{zhu2021deformable,liu2022dabdetr} and data efficiency \cite{wang2022towards}. Self-attention has also been explored in dense prediction tasks like image segmentation, where hierarchical pyramid ViT \cite{wang2021pyramidalvit}, progressive upsampling, and multi-level feature aggregation \cite{zheng2021semseg}, masking based predictions \cite{strudel2021segmenter}. In addition to these works on the image domain, Transformers have been adopted on top of convolutional feature maps for action localization and recognition \cite{Girdhar2019CVPR}, video classification \cite{wang2018nonlocal}, and group activity recognition \cite{Gavrilyuk2020groupactivity}, which were extended in pure Transformer based models considering spatio-temporal attention \cite{bertasius2021space,tong2022videomae}. In this paper, we have adapted the Transformers as part of our video encoder to consider fine-grained features and their spatial and temporal relationships to model actor-agnostic action classification tasks.
%
Besides, we take advantage of the DETR framework for multi-label action recognition, which has never been attempted in previous action models.




\textbf{Vision Language Models:} 
Numerous applications have demonstrated the high effectiveness of large-scale pretraining of image-text representations, including but not limited to, text-to-image retrieval \cite{wang2019camp}, image captioning \cite{xu2015show}, visual question answering \cite{antol2015vqa}, few and zero-shot recognition \cite{zhang2021tip, zhou2022learning}, object detection \cite{gu2021open, bangalath2022bridging, zhou2022detecting}, and image segmentation \cite{ding2022decoupling, li2022language, zhou2022extract}. As a result of their success, foundational vision-language models such as CLIP \cite{radford2021learning} and ALIGN \cite{jia2021scaling} have become quite popular in the computer vision community. However, when attempting to transfer this knowledge from vision-language models to videos, challenges arise due to the limited availability of temporal information at the image level. To address this problem, recent research such as \cite{wang2021actionclip, ni2022expanding, pan2022parameter} have attempted to adapt the popular CLIP model to videos by incorporating additional learnable components, including self-attention layers, textual or vision prompts, etc. In contrast to these existing models, our approach involves using pretrained vision-language models to create multi-modal semantic queries that can be plugged into the Transformer decoder network for extracting the key features from the video encoder for actor-agnostic multi-label classification.

\section{Multi-modal Semantic Query Network}

\begin{figure*}
\centering
\includegraphics[width=\textwidth]{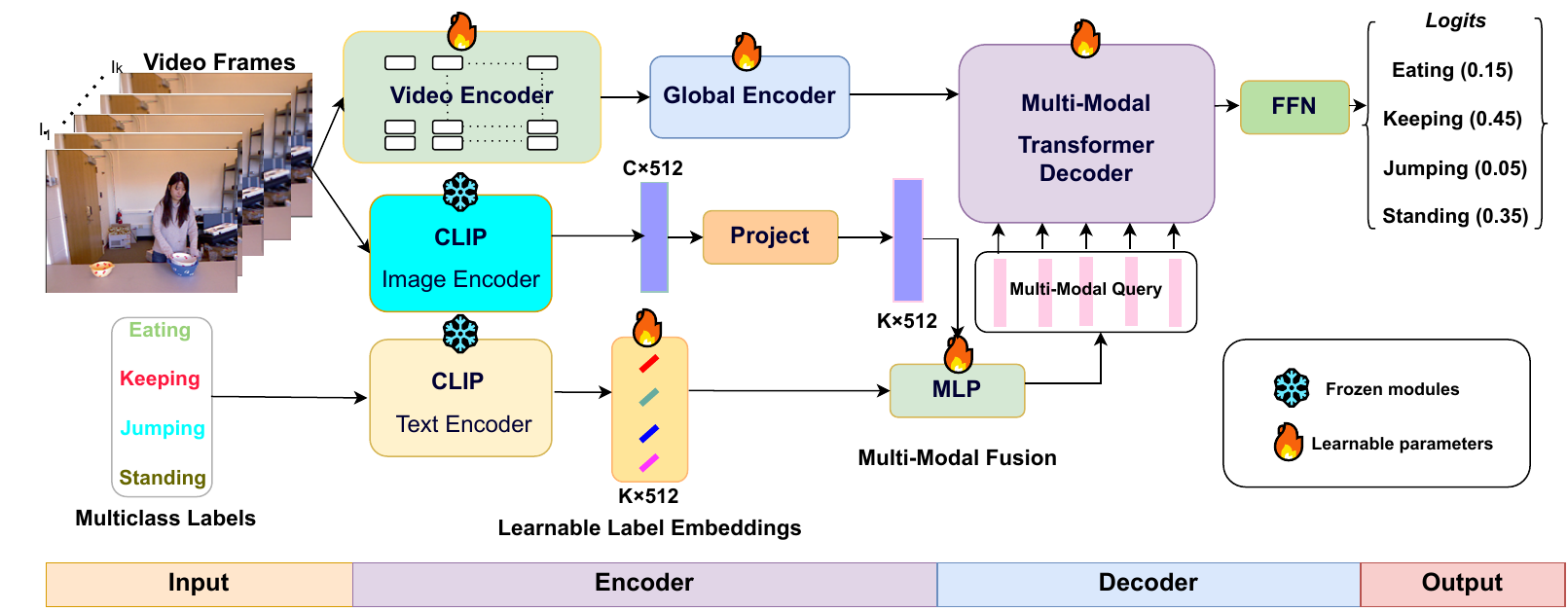}
\caption{Overview of our MSQNet for multi-modal multi-label action recognition. It has three components: a spatio-temporal video encoder, a vision-language query encoder and a multi-modal decoder. The video encoder extracts the spatio-temporal features from an input video, the query encoder merges the visual and textual information, and the multi-modal decoder transforms the video encoding to make multi-label classification with a feed-forward network (FFN).}

\label{fig:pipeline}
\end{figure*}

We introduce the MSQNet, a vision language model (VLM) in the transformer architecture, to perform multi-label multi-modal action classification in an \textit{actor-agnostic} way.
As shown in \cref{fig:pipeline},
our model consists of three components: 
(1) a \emph{spatio-temporal video encoder} that takes into account fine-grained spatial and motion cues, 
(2) a \emph{multi-modal query encoder} that combines information from both video and action class-specific sources, and
(3) a \emph{multi-modal decoder} that employs multi-headed self-attention and encoder-decoder attention mechanisms to transform the video encoding.

We start with an overview of the video encoder in \cref{sec:videnc}, followed by a description of multi-modal query encoder in \cref{sec:mmqe} and a sketch of our multi-modal decoder in \cref{sec:mmldec}. Finally, we outline our learning objective in \cref{sec:inf}.

%


\subsection{Spatio-temporal Video Encoder} \label{sec:videnc}

Consider a video $V \in \mathbb{R}^{T \times 3 \times H \times W}$ of spatial dimension $H \times W$ with $T$ sampled frames. Following the existing video Transformer models \cite{bertasius2021space,arnab2021vivit}, each frame is divided into $N$ non-overlapping square patches of size $P \times P$, with the total number of patches being $N=HW/P^2$. We flatten these patches into vectors and represented those vectors as $\vectorname{x}_{(p, t)}\in \mathbb{R}^{3P^2}$, where $p=1,\ldots,N$ denoting spatial locations and $t=1,\ldots,T$ depicting an index over frames. We then map each patch $\mathbf{x}_{(p,t)}$ into an embedding vector $\mathbf{z}_{(p,t)}^{(0)}\in\mathbb{R}^{D'}$ by a projection layer $W_\text{emb}\in\mathbb{R}^{3P^2 \times D'}$:
\begin{equation}
\mathbf{z}_{(p,t)}^{(0)}=W_\text{emb}\mathbf{x}_{(p,t)}+\mathbf{e}^{pos}_{(p,t)}
\end{equation}
where $\mathbf{e}^\text{pos}_{(p,t)}\in\mathbb{R}^{D'}$ represents a learnable positional embedding to encode the spatio-temporal position of each patch. The resulting sequence of embedded vectors $\mathbf{z}_{(p,t)}^{(0)}$ ($p=1,\ldots,N$, and $t=1, \ldots, T$) represents the input of the Transformer encoder \cite{dosovitskiy2020image}.
Following most Transformers, we add in the first position of the sequence a special learnable vector $\mathbf{z}^{(0)}_{(0,0)} \in \mathbb{R}^{D'}$ depicting the embedding of the global token. From the video encoder with $L_v$ number of layers, we thus obtain the patch level representation at each layer $l$ as:
\begin{equation}
    \mathbf{z}_{(p,t)}^{(l)}=f_{\theta_v}^{(l)}(\mathbf{z}_{(p,t)}^{(l-1)}), \hspace{10pt} l \in {1,\ldots,L_v}
\end{equation}
where $f_{\theta_v}^{(l)}$ is the $l$-th layer of the video encoder. Finally, to obtain a global frame level representation, all the patch tokens from each of the frames are averaged and then projected to a dimension $D$ using a linear projection layer (also called \emph{global encoder}) $\mathbf{W}_\text{out}\in\mathbb{R}^{D \times D'}$
\begin{equation}
\mathbf{v}_t=\mathbf{W}_\text{out}\mathbf{z}_{t}^{(L_v)}, 
\end{equation}
where $\mathbf{z}_{t}^{(L_v)}=\text{AvgPool}([\mathbf{z}_{(0,t)}^{(L_v)},\ldots,\mathbf{z}_{(N,t)}^{(L_v)}])$, $\mathbf{v}_t$ is the output representation of frame $t$ and $\mathbf{z}_{(0,0)}^{(L_v)}$ is the global token from the output sequence of the last layer of the video encoder. The sequence representing the video $V$ comprises the global token $\mathbf{z}_{(0,0)}^{(L_v)}$ and also the frame level representations $[\mathbf{v}_1, \ldots, \mathbf{v}_N]$, and takes the form of $[\mathbf{z}_{(0,0)}^{(L_v)}, \mathbf{v}_1, \ldots, \mathbf{v}_N]$, which we write as $\mathcal{F} = [\mathbf{v}_0, \mathbf{v}_1, \ldots, \mathbf{v}_N]$ by abuse of notations.

\subsection{Multi-modal Query Encoder} \label{sec:mmqe}
Given a training video $V \in \mathbb{R}^{T \times 3 \times H \times W}$ with multi-class action labels $Y$, we construct multi-modal query for our Transformer decoder network. The multi-modal query is formed by fusing the learnable label embedding and video-specific embedding. In our case, the learnable label embedding for a class is a $D$ dimensional learnable vector, depicted as $\mathcal{Q}_l\in \mathbb{R}^{K \times D}$ for a dataset, where $K$ is the total number of classes in that dataset. In our training, we initialize $\mathcal{Q}_l$ with the text embeddings of the corresponding classes. For obtaining those $D$ dimensional text embeddings, a pretrained text encoder is used, \eg, a $12$ layer CLIP \cite{radford2021learning} model (for CLIP B/16 variant) with an embedding size of $D=512$. For obtaining the video embedding, we employ the CLIP \cite{gao2021clip} image encoder (CLIP B/16 variant) on the $T$ frames independently as a batch of images and produce frame-level embeddings of dimension $D''$. These frame level embeddings are average pooled to obtain a video embedding $\mathcal{Q}_v\in\mathbb{R}^{D''}$. 
In order to form a multi-modal query, we concatenate $\mathcal{Q}_l$ and $\mathcal{Q}_v$ and deploy a linear projection with weights $\mathbf{W}_\text{que}\in\mathbb{R}^{D \times (D+D'')}$ resulting the multi-modal query $\mathcal{Q}_{0}=\mathbf{W}_\text{que}[\mathcal{Q}_{l}, \mathcal{Q}_{v}]$, where $[\cdot, \cdot]$ denotes the concatenation operation.

\myparagraph{Discussion:}
Initially, the proposed model design is independent of poses, ensuring that our solution is not limited to specific actors. Subsequently, the amalgamation of textual data and visual embeddings yields a comprehensive representation of actions, enhancing the model's expressive capacity. Lastly, incorporating general textual embedding enables the model to exhibit zero-shot capabilities.

\subsection{Multi-modal Decoder} \label{sec:mmldec}

After obtaining the spatio-temporal features $\mathcal{F}$ of input video from the video encoder, we consider the multi-modal semantic queries $\mathcal{Q}_0\in\mathbb{R}^{K\times D}$ from the multi-modal query encoder. We then perform self- and cross-attention to pool action-specific features from the spatio-temporal video representation using multi-layer Transformer decoders. We use the standard Transformer architecture, with a multi-head self-attention (MultiHeadSA) module, a cross-attention (MultiHeadCA) module, and a position-wise feed-forward network (FFN). Each decoder layer $l$ updates the queries $\mathcal{Q}_{l-1}$ from the outputs of its previous layer as follows:
\begin{align}
&\mathcal{Q}_l^{(1)}=\text{MultiHeadSA}(\tilde{\mathcal{Q}}_{l-1}, \tilde{\mathcal{Q}}_{l-1}, \mathcal{Q}_{l-1}),\\
&\mathcal{Q}_l^{(2)}=\text{MultiHeadCA}(\tilde{\mathcal{Q}}_{l}^{(1)}, \tilde{\mathcal{F}}, \mathcal{F}),\\
&\mathcal{Q}_l=\text{FFN}(\mathcal{Q}_l^{(2)}),
\end{align}
where the tilde denotes the original vectors modified by adding position encodings, $\mathcal{Q}_l^{(1)}$ and $\mathcal{Q}_l^{(2)}$ are two intermediate variables. For the sake of simplicity, we exclude the parameters of the MultiHead attention and FFN functions, which are identical to those in the standard Transformer decoder \cite{vaswani2017attention}. Each label embedding $\mathcal{Q}_{l-1, k}\in\mathbb{R}^D, k \in \{1, \ldots, K\}$, evaluates the spatio-temporal frame features $\tilde{\mathcal{F}}$ to find where to attend and then combine with the features of interest. This results in a better category-related feature for label embedding. The label embedding is then updated with this new feature. This process is repeated for each layer of the decoder network. As a result, the label embeddings $\mathcal{Q}_{k}$ are updated layer by layer and progressively injected with contextualized information from the input video via self- and cross-attention. In this way, the embeddings can be learned end-to-end from data and model label correlations implicitly.

\myparagraph{Feature projection:} To perform single-label classification, we require our model to be confident in the correct action label. For multi-label classification, we consider each predicted label as a binary classification problem. To achieve this, we project the feature representation of each class from the $L$-th layer of the Transformer decoder  $\mathcal{Q}_{L,k} \in \mathbb{R}^D$ onto a linear projection layer. This step is followed by applying an activation function $\sigma$, which is implemented as Softmax for single-label tasks and as a Sigmoid function for multi-label action classification tasks:
\begin{equation} \label{eq:projec}
p_k = \sigma(W_k \mathcal{Q}_{L,k} + b_k),
\end{equation}
where $W_k \in \mathbb{R}^D$, $\mathbf{W} = [W_1,..., W_K]^T \in \mathbb{R}^{K \times D}$, and $b_k \in \mathbb{R}$, $b = [b_1,\ldots,b_K]^T \in \mathbb{R}^K$ are the linear layer parameters and $p= [p_1,\ldots,p_K]^T \in \mathbb{R}^K$ are the probabilities of each class. We see $p$ as a function mapping an input video to class probabilities.




{
\setlength{\tabcolsep}{6pt}
\renewcommand{\arraystretch}{1} 
\begin{table*}
\centering
\resizebox{\textwidth}{!}{%
\begin{tabular}{ccclccccccc}
\hline
\multicolumn{6}{c|}{\textbf{Charades} \cite{sigurdsson2016hollywood}} &
  \multicolumn{5}{c}{\textbf{Thumos 14} \cite{idrees2017thumos}} \\ \hline
\multicolumn{1}{c|}{Method} &
  Backbone &
  \multicolumn{2}{c}{\begin{tabular}[c]{@{}c@{}}Pretrain\end{tabular}} &
  \multicolumn{1}{c|}{MMQ} &
  \multicolumn{1}{c|}{\textit{mAP}} &
  \multicolumn{1}{c|}{Method} &
  Backbone &
  \begin{tabular}[c]{@{}c@{}}Pretrain\end{tabular} &
  MMQ &
  \textit{\begin{tabular}[c]{@{}c@{}} Accuracy\end{tabular}} \\ \hline
\multicolumn{1}{c|}{AFAC \cite{zhang2021multi}} &
  Nonlocal-101 &
  \multicolumn{2}{c}{-} &
  \multicolumn{1}{c|}{No} &
  \multicolumn{1}{c|}{44.20} &
  \multicolumn{1}{c|}{SSN \cite{zhao2017temporal}} &
  C3D &
  - &
  No &
  45.42 \\
\multicolumn{1}{c|}{MViT \cite{fan2021multiscale}} &
  SlowFast &
  \multicolumn{2}{c}{K600} &
  \multicolumn{1}{c|}{No} &
  \multicolumn{1}{c|}{43.90} &
  \multicolumn{1}{c|}{R-C3D \cite{xu2019two}} &
  C3D &
  - &
  No &
  57.19 \\
\multicolumn{1}{c|}{ActionCLIP \cite{wang2021actionclip}} &
  ViT-B &
  \multicolumn{2}{c}{-} &
  \multicolumn{1}{c|}{No} &
  \multicolumn{1}{c|}{44.30} &
  \multicolumn{1}{c|}{BMN \cite{lin2019bmn}} &
  C3D &
  S1M &
  No &
  62.12 \\ \hline
\multicolumn{1}{c|}{{MSQNet}} &
  ViT-B &
  \multicolumn{2}{c}{K400} &
  \multicolumn{1}{c|}{No} &
  \multicolumn{1}{c|}{43.99} &
  \multicolumn{1}{c|}{{MSQNet}} &
  C3D &
  K400 &
  No &
  67.47 \\
\multicolumn{1}{c|}{{MSQNet}} &
  {TS} &
  \multicolumn{2}{c}{{K400}} &
  \multicolumn{1}{c|}{{No}} &
  \multicolumn{1}{c|}{{44.11}} &
  \multicolumn{1}{c|}{\textbf{MSQNet}} &
  {{TS}} &
  {{K400}} &
  {{Yes}} &
  \textbf{79.71} \\ 
\multicolumn{1}{c|}{\textbf{MSQNet}} &
  {TS} &
  \multicolumn{2}{c}{{K400}} &
  \multicolumn{1}{c|}{{Yes}} &
  \multicolumn{1}{c|}{\textbf{47.57}} &
  \multicolumn{1}{c|}{\textbf{MSQNet}} &
  {{TS}} &
  {{K400}} &
  {{Yes}} &
  \textbf{83.16} \\ \hline
  \hline
\multicolumn{6}{c|}{\textbf{Animal Kingdom} \cite{ng2022animal}} &
  \multicolumn{5}{c}{\textbf{Hockey} \cite{sozykin2018multi}} \\ \hline
\multicolumn{1}{c|}{Method} &
  Backbone &
  \multicolumn{2}{c}{\begin{tabular}[c]{@{}c@{}}Pretrain\end{tabular}} &
  \multicolumn{1}{c|}{MMQ} &
  \multicolumn{1}{c|}{\textit{mAP}} &
  \multicolumn{1}{c|}{Method} &
  Backbone &
  \begin{tabular}[c]{@{}c@{}}Pretrain \end{tabular} &
  MMQ &
  \textit{\begin{tabular}[c]{@{}c@{}}Multilabel \\ Accuracy\end{tabular}} \\ \hline
\multicolumn{1}{c|}{CARe \cite{ng2022animal}} &
  X3D &
  \multicolumn{2}{c}{-} &
  \multicolumn{1}{c|}{No} &
  \multicolumn{1}{c|}{25.25} &
  \multicolumn{1}{c|}{EO-SVM \cite{carbonneau2017multiple}} &
  - &
  - &
  No &
  90.00 \\
\multicolumn{1}{c|}{CARe \cite{ng2022animal}} &
  I3D &
  \multicolumn{2}{c}{-} &
  \multicolumn{1}{c|}{No} &
  \multicolumn{1}{c|}{16.48} &
  \multicolumn{1}{c|}{AFAC \cite{zhang2021multi}} &
  CSN-152 \cite{tran2019video} &
  - &
  No &
  96.30 \\ \hline
\multicolumn{1}{c|}{\textbf{MSQNet}} &
  {I3D} &
  \multicolumn{2}{c}{{-}} &
  \multicolumn{1}{c|}{No} &
  \multicolumn{1}{c|}{\textbf{55.59}} &
  \multicolumn{1}{c|}{{MSQNet}} &
  {I3D} &
  {-} &
  No &
  93.29 \\
\multicolumn{1}{c|}{\textbf{MSQNet}} &
  {{TS}} &
  \multicolumn{2}{c}{{{K400}}} &
  \multicolumn{1}{c|}{{No}} &
  \multicolumn{1}{c|}{\textbf{71.63}} &
  \multicolumn{1}{c|}{\textbf{MSQNet}} &
  {TS} &
  {K400} &
  {No} &
  {93.45} \\ 
\multicolumn{1}{c|}{\textbf{MSQNet}} &
  {{TS}} &
  \multicolumn{2}{c}{{{K400}}} &
  \multicolumn{1}{c|}{{Yes}} &
  \multicolumn{1}{c|}{\textbf{73.10}} &
  \multicolumn{1}{c|}{\textbf{MSQNet}} &
  {TS} &
  {K400} &
  {Yes} &
  \textbf{96.95} \\ \hline
\end{tabular}%
}
\caption{Comparing our MSQNet against the state-of-the-art in the supervised learning setting. The best results are in bold.
\texttt{K400}: Kinetics-400;
\texttt{K600}: Kinetics-600;
\texttt{S1M}: Sports-1M;
\texttt{TS}: TimeSformer;
\texttt{MMQ}: Multi-modal Query.}
\label{tab:fsscores}
\end{table*}
}

\subsection{Learning Objective} \label{sec:inf}
Given a video $V$, our objective is to train our model in a way so that the predicted probability for each action class $p = [p_1,\ldots,p_K]^T \in \mathbb{R}^K$ matches with the ground truth. Thus we train our model using the categorical cross-entropy loss as the final learning objective. 
\begin{align}
\mathcal{L} = -\frac{1}{n}\sum_{i=1}^n\sum_{j=1}^K y_{ij}\log(p_{ij}), 
\end{align}
where $n$ is the number of samples or observations and $K$ is the number of classes. Specifically, on a single-label dataset, such as Thumos14, we use the ``classical'' cross-entropy loss useful for multi-class single-label settings, whereas, for other datasets which are multi-label, we use binary cross-entropy loss. 

%
%

\section{Experiments} 
\label{exp}

\myparagraph{Datasets:} 
We evaluate both single-label and multi-label action recognition datasets: (1) \heading{Thumos14} \cite{idrees2017thumos}, a single-label dataset, consists of 13,000 videos from 20 classes, with 1,010 validation and 1500 untrimmed test videos. (2) \heading{Hockey} \cite{sozykin2018multi}, a multi-label dataset, has 12 activities across 36 videos. (3) \heading{Charades} \cite{sigurdsson2016hollywood} contains 66,500 annotations for 157 actions and is divided into 7,986 training and 1,863 validation multi-label videos. (4) \heading{Animal Kingdom} \cite{ng2022animal} is a large multi-label dataset with over 50 hours of footage featuring wild animals from different classes and environments. It consists of 30,000 video sequences, including over 850 species from Mammalia, Aves, Reptilia, Amphibia, Pisces, and Insects. (5) \heading{HMDB51} \cite{kuehne2011hmdb} is a comprehensive compilation of original videos from various sources, including films and online videos. It comprises 6,766 video clips spanning 51 different action categories, such as “jump”, “kiss”, “laugh” etc. Each class consists of at least 101 clips.

\myparagraph{Training details:} 
We train our model for $100$ epochs with a cosine decay scheduler and an initial learning rate of $ 0.00001$ using Adam optimizer \cite{Diederik2015Adam}. Unless otherwise stated, we set the number of frames for training to $16$. We use the BCEWithLogitsLoss for Animal Kingdom, Charades Hockey, Volleyball, and the CrossEntropyLoss for Thumos14 and HMDB51. 

\myparagraph{Evaluation metrics:} 
Following the existing protocol, we use accuracy  \cite{Nicki_Skafte_Detlefsen_and_Jiri_Borovec_and_Justus_Schock_and_Ananya_Harsh_and_Teddy_Koker_and_Luca_Di_Liello_and_Daniel_Stancl_and_Changsheng_Quan_and_Maxim_Grechkin_and_William_Falcon_TorchMetrics_-_Measuring_2022} as the evaluation metric for Hockey, Volleyball, Thumos14 and HMDB51. For Animal Kingdom and Charades datasets, we use the mean average precision (mAP) \cite{Su2015PerfEvalIR} for performance measurement.

\begin{table}[!h]
\centering

\resizebox{\columnwidth}{!}{
\begin{tabular}{ccccc}
\hline
Method          & Backbone & Pretrain & MM & Accuracy       \\ \hline
BIKE \cite{wu2022bidirectional}         & ViT        & WIT-400M        & Yes   & 84.31              \\
	
R2+1D-BERT \cite{kalfaoglu2020late}          & R(2+1)D         & IG65M        & No   &    85.10           \\
VideoMAE V2-g \cite{wang2023videomae}          & ViT        & K400/K600        & No   & 88.10              \\ \hline
\textbf{MSQNet} & TS       & K400     & Yes & \textbf{93.25} \\ \hline
\end{tabular}
}
\caption{Supervised learning performance on HMDB51. The best results are highlighted in bold.}
\label{tab:hmdbfs}
\end{table}

\subsection{Supervised learning evaluation}

\myparagraph{Setting:} This is the most conventional setting where a labeled dataset $D_S$ with labels $Y_S=\lbrace{Y_i\rbrace}_{i=1}^{n}$ is available for model training.

\myparagraph{Results:} 
We present the comparative results
in \cref{tab:fsscores}. 
For the Charades dataset, we compare our model with AFAC \cite{zhang2021multi}, MViT \cite{fan2021multiscale}, and ActionCLIP \cite{wang2021actionclip}. 
In the case of the Thumos14 dataset, we have considered BMN \cite{lin2019bmn}, R-C3D \cite{xu2019two}, and SSN \cite{zhao2017temporal}. For Animal Kingdom, we only consider CARe \cite{ng2022animal} with two backbones (X3D and I3D),
as no other method presented their results on it. For the Hockey dataset, we consider EO-SVM \cite{carbonneau2017multiple} and AFAC \cite{zhang2021multi}. Finally, for HMDB51 in \cref{tab:hmdbfs}, we consider BIKE \cite{wu2022bidirectional}, R2+1D-BERT \cite{kalfaoglu2020late}, and VideoMAE V2-g \cite{wang2023videomae} for comparison. We have used two settings to evaluate MSQNet, one with learnable queries (\ie~only text cues, without the video cues) and the other with multi-modal queries, considering both the learnable queries and video cues. We utilized the backbone of ViT-B, C3D, and TimeSformer trained on the Kinetics-400 \cite{kay2017kinetics} and I3D for fair comparison. As seen in \cref{tab:fsscores} and \cref{tab:hmdbfs}, our multi-modal setting surpasses all the previous approaches. Notably, the improvement is significant for the Animal Kingdom \cite{ng2022animal} dataset, suggesting that integrating vision-language information is beneficial for handling diverse actions across various species and genera. Furthermore, MSQNet's outstanding performance on the three human action datasets confirms the generalizability of our model, as it is independent of specific actors. 

\subsection{Zero-shot learning evaluation}

\myparagraph{Setting:} \label{para:zss} In this setting, the model is trained on a source dataset $D_\text{train}$ and tested directly on a target dataset $D_\text{test}$. The source dataset $D_\text{train}$ contains samples belonging to source classes $Y_\text{train}=\lbrace{y_i\rbrace}_{i=0}^{k}$. The model is evaluated on the target dataset $D_\text{test}$ with classes $Y_\text{test}$, such that $Y_\text{train} \bigcap Y_\text{test} = \phi$, \ie, the action categories for training and testing remain disjoint. For this experiment, we have considered three datasets: (1) Thumos14 \cite{idrees2017thumos}, (2) Charades \cite{sigurdsson2016hollywood}, and (3) HMDB51 \cite{kuehne2011hmdb}. For Thumos14, we consider the dataset splits proposed by \cite{nag2022zero}. For HMDB51, we consider the zero-shot splits by \cite{qian2022multimodal}. In the case of Charades \cite{sigurdsson2016hollywood}, since no such split is publicly available, we have defined our own random splits for conducting the zero-shot experiments, which we will make publicly available upon the acceptance of this paper. Specifically, for this experimental setting, we consider two different dataset splits for Thumos14, Charades, and HMDB51: (1) $Y_\text{train}$ and $Y_\text{test}$ respectively contain 75\% and 25\% of the total number of classes, (2) $Y_\text{train}$ and $Y_\text{test}$ respectively contain 50\% and 50\% of the total number of classes in the dataset. To ensure statistical significance, we have followed \cite{nag2022zero} and considered 10 different random splits of action classes with the aforementioned settings. 

\myparagraph{Results:} We present the experimental results for zero-shot setting in \cref{zs}.
Unfortunately, we were unable to run VideoCOCA \cite{yan2022video}, CLIP-Hitchiker \cite{bain2022clip} and, BIKE \cite{wu2022bidirectional} on our splits for the unavailability of their open-source codes. Therefore, we followed the settings mentioned in  \cref{para:zss} to ensure a fair evaluation, and we still report their scores for all the datasets. While comparing, we consider different components of our MSQNet model in \cref{zs}: (1) \heading{Vanilla MSQNet:} MSQNet without text initialization and video embeddings; (2) \heading{Vanilla MSQNet + Text Init.:} MSQNet with text initialisation, but without video embeddings; (3) \heading{MSQNet:} our full model with text initialisation and video embeddings. We further compare MSQNet against their respective SoTA for all three datasets. The results in \cref{zs} demonstrate that our MSQNet model outperforms the baselines by a significant margin, highlighting the effectiveness of the different model components.  Furthermore, the results emphasize the importance of video embeddings for achieving zero-shot capabilities in the two datasets.

\begin{table}[!t]
\centering
\resizebox{\columnwidth}{!}{
\begin{tabular}{c|cccc}
\hline
Split                      & Method                              & \textbf{Thumos 14} & \textbf{Charades} & \textbf{HMDB51} \\ \hline
\multirow{3}{*}{Reported}  & VideoCOCA \cite{yan2022video}       & -                  & 21.1              & 58.70           \\
                           & CLIP-Hitchhiker \cite{bain2022clip} & -                  & 25.8              & -               \\
                           & BIKE                                & -                  & -                 & 61.40           \\ \hline
\multirow{3}{*}{50\% Seen} & Vanilla MSQNet                      & 49.37              & 15.87             & 45.66           \\
                           & Vanilla MSQNet + Text Init          & 53.76              & 18.30             & 51.22           \\
                           & \textbf{MSQNet }                    & \textbf{63.98}     & \textbf{30.91}    & \textbf{59.24}  \\ \hline
\multirow{3}{*}{75\% Seen} & Vanilla MSQNet                      & 52.02              & 17.43             & 48.37           \\
                           & Vanilla MSQNet + Text Init          & 60.28              & 18.62             & 59.58           \\
                           & \textbf{MSQNet}                     & \textbf{75.33}     & \textbf{35.59}    & \textbf{69.43}  \\ \hline
\end{tabular}%
}
\caption{Comparing MSQNet against state-of-the-art in zero-shot setting. We report multilabel accuracy scores for Thumos 14, HMDB51, and mAP for the Charades dataset. The best results are in bold.}
\label{zs}
\end{table}

\begin{table}[!h]
\centering
\begin{tabular}{c|cc}
\hline
\textbf{Backbone} & \begin{tabular}[c]{@{}c@{}}\textbf{Animal Kingdom}\\ (mAP)\end{tabular} & \begin{tabular}[c]{@{}c@{}}\textbf{Charades}\\ (mAP)\end{tabular} \\ \hline
VideoMAE \cite{tong2022videomae} & 71.19 & 41.69 \\
TimeSformer \cite{bertasius2021space} & \textbf{73.10} & \textbf{47.57} \\ \hline
\end{tabular}
\caption{Performance of MSQNet using different weights for the video encoder. The best scores are in bold.}
\label{vid_bb}
\end{table}

\begin{table}[!h]
\centering
\begin{tabular}{c|cc}
\hline
\textbf{\# Frames} & \textbf{AK} (mAP) & \textbf{Charades} (mAP)\\ \hline
\multicolumn{1}{c|}{{8}} & \textcolor{black}{67.74} & {43.06} \\
\multicolumn{1}{c|}{{10}} & \textbf{\textcolor{blue}{69.73}} & \textbf{\textcolor{blue}{45.33}} \\
\multicolumn{1}{c|}{{16}} & \textbf{\textcolor{red}{73.10}} & \textcolor{red}{\textbf{47.57}} \\ \hline
\end{tabular}
\caption{Performance of MSQNet with a different number of frames. The best and the second-best scores are in red and blue. AK: Animal Kingdom}
\label{diff_frames}
\end{table}

\begin{figure*}[!t]
\centering
\includegraphics[width=\textwidth]{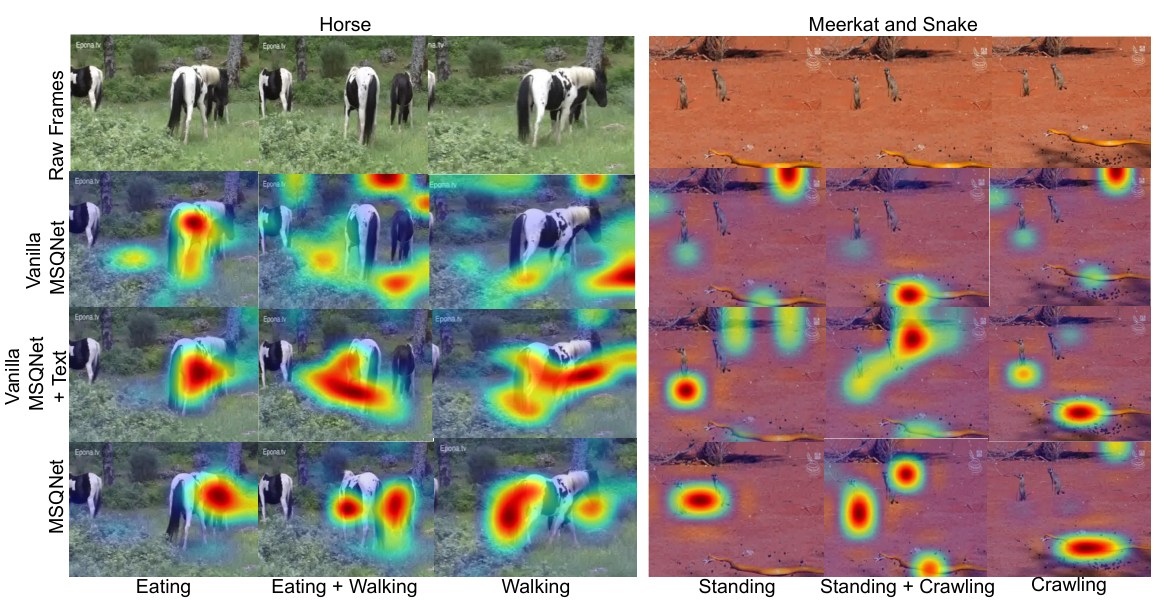}
\caption{Attention rollout on sample videos from Animal Kingdom \cite{ng2022animal} showing raw frames, heatmap with only bare backbone, with uni-modal prompt, and MSQNet.}
\label{fig:gc}
\end{figure*}

\subsection{Further analysis}

\myparagraph{Video encoder:} In \cref{vid_bb}, we demonstrate the impact of various backbones on the performance of MSQNet. Our analysis focuses on state-of-the-art backbone architectures such as VideoMAE \cite{tong2022videomae} and TimeSformer \cite{bertasius2021space}, which we use to initialize our video encoder. We leverage the Animal Kingdom and Charades datasets to assess the efficacy of these backbones. Our results reveal that the TimeSformer \cite{bertasius2021space} backbone outperforms the VideoMAE \cite{tong2022videomae} backbone on both datasets. The success of TimeSformer can be attributed to its ``divided attention'', which enables the network to attend separately to spatial and temporal features within each block, leading to enhanced video classification accuracy.

\myparagraph{Number of frames}: In our MSQNet model, we have considered videos with $16$ (\ie, $T=16$) frames as a default setting. However, we have also experimented with clips of lengths 8 and 10 frames. As we can see from \cref{diff_frames}, using 16 frames gives us the best performance. This finding is attributed to the fact that sampling more frames provides a more extensive comprehension of intricate actions and events that can occur over an extended duration. Our finding is consistent over both human and animal action datasets, respectively.



\myparagraph{Importance of multi-modal queries:} 
The results presented in the final two rows for each dataset in \cref{tab:fsscores} exemplify the advantage of integrating learnable queries (\ie~textual features) and visual features to augment the performance of our model. Through visual and textual cues, our model can achieve a more comprehensive understanding of the context surrounding a particular scene or action. This is particularly beneficial in animal datasets, where the CLIP image encoder can be employed to its full potential. This is because the CLIP has undergone pretraining on millions of image-text pairs \cite{radford2021learning}, thereby gaining outstanding zero-shot capability. Consequently, our model excels in diverse scenarios, accommodating a wide range of actors. 







\begin{figure*}[!ht]
\centering
\includegraphics[width=\textwidth]{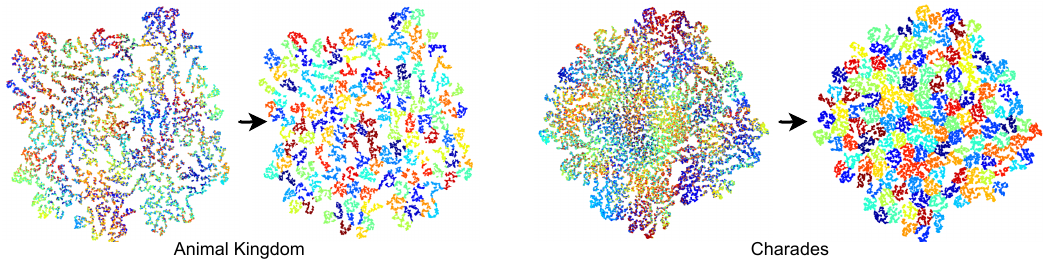}
\caption{Video embeddings without and with the proposed multi-modal query learning  on Animal Kingdom and Charades. Arrow shows the transition.}
\label{fig:tsne}
\end{figure*}

\begin{figure}[!ht]
\centering
\includegraphics[width=\columnwidth]{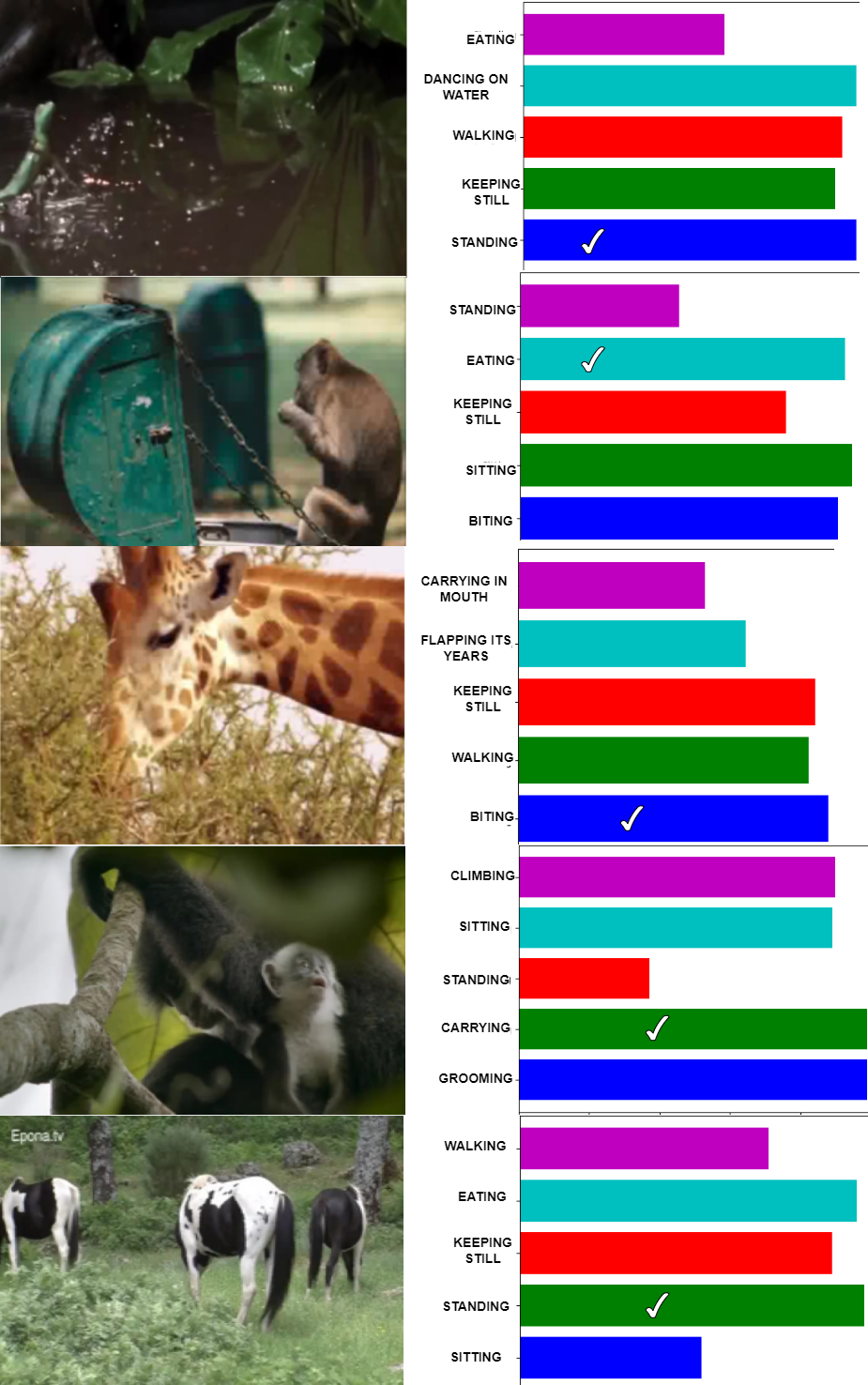}
\caption{Confidence scores of top-5 classes predicted by our MSQNet. Correctly classified action classes are marked with a \checkmark.} 
\label{fig:moreviz}
\end{figure}

\myparagraph{Effect of Text Encoding:} We evaluate the effect of text encoding in the supervised setting of our MSQNet model. We have tested two common text encoders: BERT \cite{devlin2018bert} and CLIP \cite{radford2021learning}. In \cref{tab:texten}, we observe that the multi-modality (vision and language) learning based on CLIP is superior to the pure language model BERT. This is not surprising that the former boasts a more robust feature embedding capability, given its extensive pretraining with millions of image-text pairs. Further, it is noteworthy that their difference is not substantial, suggesting the robustness of our method in the text encoding component.

\begin{table}[!h]
\centering
\begin{tabular}{c|cc}
\hline
\textbf{Text Encoder} & \textbf{AK} (mAP) & \textbf{Charades} (mAP)\\ \hline
BERT \cite{devlin2018bert} & 69.81 & 43.57 \\
CLIP \cite{radford2021learning} & \textbf{71.63} & \textbf{44.11}  \\ \hline
\end{tabular}
\caption{Effect of text encoding with our MSQNet.}
\label{tab:texten}
\end{table}

\subsection{Visualization}

\myparagraph{Qualitative analysis:} We employed a feature attention visualization to examine the behavior of MSQNet, as shown in \cref{fig:gc}. In row 2, we observed that the Vanilla MSQNet (the one with the video encoder alone concentrates on the background to classify the ``Eating'' and ``Walking'' actions for the ``Horse'' diagram. However, when we introduced textual information to the Vanilla MSQNet (row 3), the attention shifted entirely to the horse's body. We enhanced the model's performance by integrating the CLIP image encoder into the uni-modal MSQNet. The attention heatmap demonstrates that introducing a multi-modal prompt to MSQNet allows it to accurately classify ``Eating'' by emphasizing key features such as the horse, mouth, and grass. Meanwhile, when predicting ``Walking'' (row 4), the attention focuses on the horses' legs. Similarly, for the ``Meerkat and Snake'' diagram, MSQNet (row 4) correctly attends to the intended frame regions while identifying the actions ``Standing'', ``Standing and Crawling'', and ``Crawling'' more effectively than their uni-modal and bare video backbone counterparts. This confirms the effectiveness of our approach of leveraging pre-trained vision-language knowledge for accurate action classification. 

Further, inspired by \cite{feichtenhofer2017spatiotemporal}, we visualize the confidence scores for the top-5 classes predicted by MSQNet, as shown in \cref{fig:moreviz} and \href{https://i.imgur.com/GPoqH8C.gif}{https://i.imgur.com/GPoqH8C.gif}. Notably, all the top-class predictions exhibit a strong correlation, indicating the robust generalization capabilities of our model.

\myparagraph{Feature visualization :} The plots in \cref{fig:tsne} show the t-SNE \cite{van2008visualizing} diagram of the video representations from the Animal Kingdom and Charades datasets. The visualizations demonstrate that the embeddings of the action classes become more distinguishable and meaningful after passing through the multi-modal transformer decoder. 
These findings suggest that the MSQNet is capable of accurately classifying actions, regardless of the characteristics of the datasets being used.


\section{Conclusion}
Our proposed MSQNet model utilizes visual and textual information from a pretrained vision-language model to accurately define action classes, eliminating the need for actor-specific design. We achieve improved model design and maintenance efficiency by framing the problem as a multi-modal target detection task within the Transformer decoder. Extensive experiments conducted on multiple benchmarks demonstrate the superiority of our approach over previous actor-specific alternatives for multi-label action recognition tasks involving both humans and animals as actors, including both fully-supervised and zero-shot scenarios.

In the future, we have plans to explore advancements in multi-modal learning techniques further and explore the integration of additional modalities, such as audio, to enhance our model's capabilities. Furthermore, we aim to extend our model to address action detection tasks, allowing for a more comprehensive video action understanding.

\clearpage

{\small
\bibliographystyle{ieee_fullname}
\bibliography{egbib}
}

\end{document}